\documentclass[sigconf,nonacm, natbib=true]{acmart}

\usepackage{float}
\usepackage{array,graphicx}
\usepackage{siunitx}
\usepackage{caption}
\usepackage{subcaption}

\AtBeginDocument{
  \providecommand\BibTeX{{
    \normalfont B\kern-0.5em{\scshape i\kern-0.25em b}\kern-0.8em\TeX}}}

\usepackage{multirow}

\begin{document}

\title{Performance Prediction for Multi-hop Questions}

\author{Mohammadreza Samadi}
\email{samadi2@ualberta.ca}
\affiliation{
  \institution{University of Alberta}
  \city{Edmonton}
  \state{Alberta}
  \country{Canada}
}

\author{Davood Rafiei}
\email{drafiei@ualberta.ca}
\affiliation{
  \institution{University of Alberta}
  \city{Edmonton}
  \state{Alberta}
  \country{Canada}
}

\begin{abstract}
We study the problem of Query Performance Prediction (QPP) for open-domain multi-hop Question Answering (QA), where the task is to estimate the difficulty of evaluating a multi-hop question over a corpus. Despite the extensive research on predicting the performance of ad-hoc and QA retrieval models, there has been a lack of study on the estimation of the difficulty of multi-hop questions. The problem is challenging due to the multi-step nature of the retrieval process, potential dependency of the steps and the reasoning involved. To tackle this challenge, we propose multHP, a novel pre-retrieval method for predicting the performance of open-domain multi-hop questions. Our extensive evaluation on the largest multi-hop QA dataset using several modern QA systems shows that the proposed model is a strong predictor of the performance, outperforming traditional single-hop QPP models. Additionally, we demonstrate that our approach can be effectively used to optimize the parameters of QA systems, such as the number of documents to be retrieved, resulting in improved overall retrieval performance.

\end{abstract}

\maketitle

\section{Introduction}
The task of open-domain QA--answering questions over a massive collection of documents--has received much attention lately due to the rise of conversational assistant systems such as Apple Siri, Amazon Alexa, Microsoft Cortana, and Google Assistant.
Traditional IR models (e.g., BM25~\cite{robertson1995okapi}) have been particularly popular as retrievers in this task \cite{chen-etal-2017-reading, wang2018r,wang2019multi,yang2019end} thanks to their simplicity and fast response time.
In particular, the \textit{retrieve-and-read} framework consists of two components, a retriever and a reader, where the retriever extracts relevant documents from a large collection of documents, and the reader aggregates information in the retrieved documents and extracts the answer \cite{chen-etal-2017-reading}.
Our work is focused on open-domain multi-hop QA where one has to reason with the information that is spread over more than one document in the collection to reach an answer.
For example, consider the question ``In what year did the young actor who co-starred with Sidney Poitier in Little Nikita die, and what was the cause of death?''. To answer this question, one may first retrieve a document that mentions both Sidney Poitier and Little Nikita. This can be, for example, a document that lists the cast members of Little Nikita. From this document, one may find out that the question is referring to River Phoenix. The document about Little Nikita is less likely to give more information about River Phoenix though.
Hence in another retrieval step, the Wikipedia document of River Phoenix may be retrieved to find out that the actor died in 1993 due to drug intoxication.
This is one form of multi-hop, referred to as a \textit{bridge question}. Another form of multi-hop includes questions that must compare pieces of information or reason over facts spread in multiple documents. For instance, the question ``Were Stanley Kubrick and Elio Petri from different countries?'' is a \textit{comparison question}. First, both documents about Stanley Kubrick and Elio Petri will be retrieved, and after a reasoning process that compares their nationalities, the question can be answered.

The specific question we investigate in this paper is if
the performance of a multi-hop question can be predicted and if such a prediction is a good proxy for the performance of multi-hop QA systems, some of which are more complex.
Query Performance Prediction (QPP) plays an important role in resource allocation and evaluating the performance of different systems. For example, QPP can assist a search engine in allocating additional resources to more difficult questions.
Even though QPP has been extensively studied,
with methods ranging from statistical approaches \cite{hauff2008improved, mothe2005linguistic,perez2010standard,shtok2012predicting} to more recent neural models \cite{hashemi2019performance, arabzadeh2021bert},
we are not aware of any such work on multi-hop questions.
The models that utilize the retrieve-and-read framework \cite{chen-etal-2017-reading} in open-domain QA fail in multi-hop questions for a few reasons. First, the clues for answering the questions are often spread over multiple supporting documents, and all supporting documents may not be retrieved in one step of the retrieval, as shown in our Little Nikita example. Hence an iterative retrieve-and-read process may be needed \cite{qi2019answering}.
Second, many questions require some form of reasoning over the facts described in supporting documents, and standard retrievers are oblivious to such a chain of reasoning when the facts are spread in multiple documents.

Our approach utilizes retrieval paths to decompose each question into a few retrieval steps, starting from the question and ending with a document that has an answer.
We show that retrieval path types can be detected from questions and that the performance of a multi-hop question can be estimated based on that of its retrieval steps, for which corpus-based statistics are more likely to be available and standard QPP models may be applicable.

Our contributions can be summarized as follows:
\begin{itemize}
\item We introduce retrieval paths as a model of the retrieval steps a QA system is expected to take, starting from a multi-hop question and ending with an answer.
\item We develop a performance prediction model that utilizes retrieval paths to estimate how much a multi-hop question can be difficult to answer.
\item Our evaluation reveals that the proposed model is a strong predictor of the actual performance of a few QA systems (including some dense models) while outperforming the relevant QPP baselines from the literature.
\item We study two use cases of our performance prediction, showing that our model can be effectively used for annotating QA datasets and within an adaptive retriever. 

\end{itemize}

The rest of the paper is organized as follows: Section \ref{sec:related_work} reviews related work on Multi-hop QA and Query Performance Prediction. Section \ref{sec:methodology} discusses our methodology for defining retrieval paths and estimating difficulty scores. Section \ref{sec:results} outlines the experimental setup and discusses our results. Finally, Section \ref{sec:conclusions} concludes this paper and discusses potential future works.
 
\section{Related Work}
\label{sec:related_work}
Our work is related to the lines of work on open-domain multi-hop question answering and query performance prediction.

\subsection{Open-Domain Multi-hop Question Answering}
The release of large-scale datasets for multi-hop QA \cite{welbl2018constructing, yang-etal-2018-hotpotqa, ho-etal-2020-constructing, trivedi-etal-2022-musique} has invigorated the research interest in this area. 

The studies predominantly utilize an iterative \textit{retrieve-and-read} framework to secure the supporting documents, where the performance is closely linked to that of the retriever (see \citet{mavi2022survey} for a survey). The work in this area may be categorized into iterative retrieval and entity-based retrieval.

\subsubsection{Iterative retrievers}
An essential step in iterative retrievers is extracting clues such as bridge entities or other relevant information from the documents retrieved for the current hop and updating the query for the next hop~\citep{xiong2019simple,zhang2021answering}.
\citet{qi2019answering} propose GoldEn, an iterative \textit{retrieve-and-read} approach  that uses both the question and the retrieved documents at each step to generate a query for the next step. The query generation is done using a supervised model that is trained on the semantic overlap between retrieved contexts and the documents to be retrieved. This approach is shown to perform well on HotpotQA when integrated with DrQA reader \cite{chen-etal-2017-reading}.

For better capturing the query semantics, \citet{xiong2020answering} introduce a dense retriever, referred to as MDR, that chooses a supporting document in each iteration based on the cosine similarity of the query dense vector and the document vectors. The question is updated at each step by concatenating the question vector with the vector of a retrieved document. 
\citet{feldman2019multi} use a combination of sparse and dense retrievers whereas \citet{wu2022triple} represent each document with a set of triple facts in the form of <subject, predicate, object>, which are used to update the question. There are also works that adapt to the number of hops in the iterative process~\citep{asai2020learning,zhu2021adaptive}.
Despite using lexical features, our QPP model is evaluated in the context of both sparse and dense retrievers.

\subsubsection{Using entities in retrievals}
The main challenge in retrieving the supporting documents of a multi-hop question is missing entities, and that the question text  may not be enough to find all relevant documents. Several studies utilize the relationship between different entities, including those explicitly mentioned and those missing, to retrieve relevant evidence~\citep{das2019multi,shao2021memory,zhaotransformer,li2021hopretriever}. For example, \citet{ding-etal-2019-cognitive} construct a graph from entities in the question and supporting documents and train a model to predict the most promising entity for the next hop. Similar to these approaches, entities mentioned in questions are also used within our performance prediction model. 

\subsection{Query Performance Prediction}
\subsubsection{Query performance prediction in IR}
This line of work can be categorized into pre-retrieval and post-retrieval. 
Our work fall under pre-retrieval QPP, where a prediction is performed without performing a retrieval and solely based on the query expression and corpus statistics. Pre-retrieval methods may estimate the query performance based on the similarity between the query and the document collection~\citep{zhao2008preretrieval}, on the basis that queries that are more similar to the collection are easier.
Some approaches use
the pair-wise similarity or coherence of documents that contain the query terms ~\citep{he2008using,zhao2008preretrieval} whereas others use the frequency or the specificity of query terms~\citep{cronen2002predicting,he2004inferring} to estimate the performance.
Since frequency-based methods ignore the semantic relationships between non-matching terms, some works use the cluster of words around a term to estimate the specificity~\citep{roy2019estimating,arabzadeh2020neural}.

Being a pre-retrieval method, our approach also uses query terms and corpus statistics in the estimation process but it differs from previous specificity-based approaches in two important aspects. First, our approach does the estimation in the context of retrieval paths and question types, allowing us to provide more accurate estimation for each retrieval path. Second, given the nature of questions in open-domain settings, we aim at using in our estimation salient question terms that play a role in an effective retrieval, rather than considering all question terms.

\subsubsection{Query performance prediction in QA}
QPP is studied in the context of QA due to the importance of retrieval systems in open-domain QA pipelines.
\citet{krikon2012predicting} decompose the effectiveness of a set of passages retrieved to a question into (1) the probability that the information need of the question is satisfied by the passages, and (2) the probability that the passages contain the answers.
The former is estimated using post-retrieval QPP models such as the clarity score~\cite{cronen2002predicting} and other metrics~\cite{zhou2007query, shtok2012predicting}, 
whereas the latter is judged based on the presence of named entities that may answer the question.
To address the problem of QPP for non-factoid questions, \citet{hashemi2019performance} develop a neural model that has three components: (1) the scores assigned to candidate answers by the QA system, (2) query performance, estimated using pre-retrieval QPP models, and (3) the content of top \textit{k} retrieved answers.

These approaches all fall under post-retrieval methods, wherein ranked results are retrieved to predict a difficulty score. We are not aware of pre-retrieval QPP models being studied in the context of Open-domain QA prior to our work.

\begin{figure*}[tb]
\centering
\begin{subfigure}{0.36\textwidth}
  \centering\includegraphics[width=\textwidth]{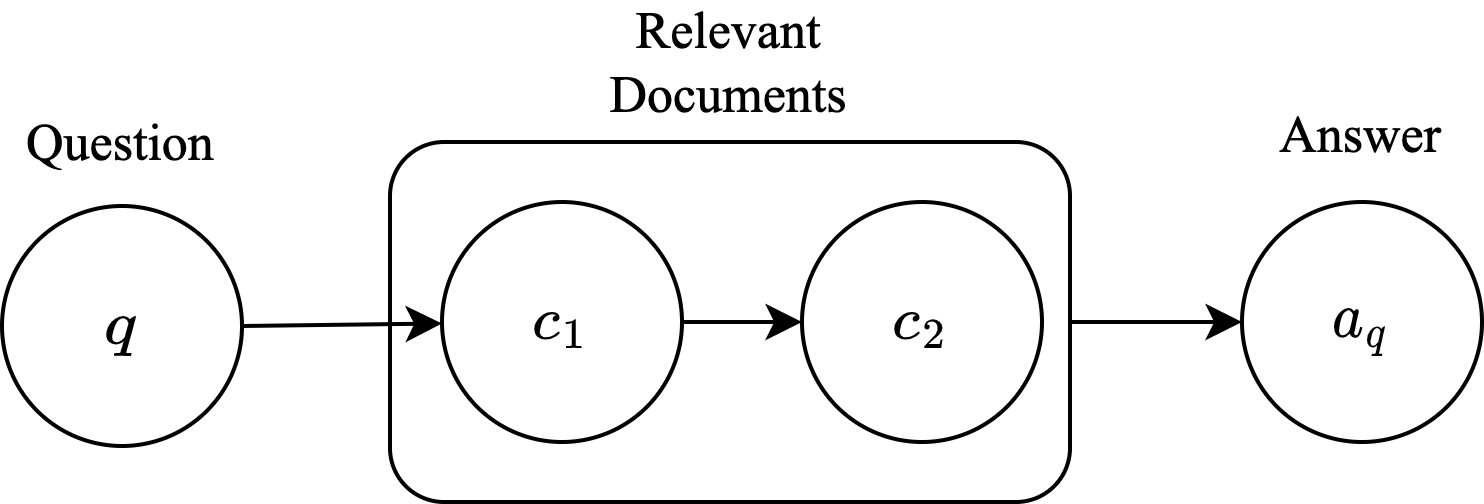}
  \caption{Bridge}
  \label{fig:rpath-bridge}
\end{subfigure}
\hfill
\begin{subfigure}{0.27\textwidth}
  \centering\includegraphics[width=\textwidth]{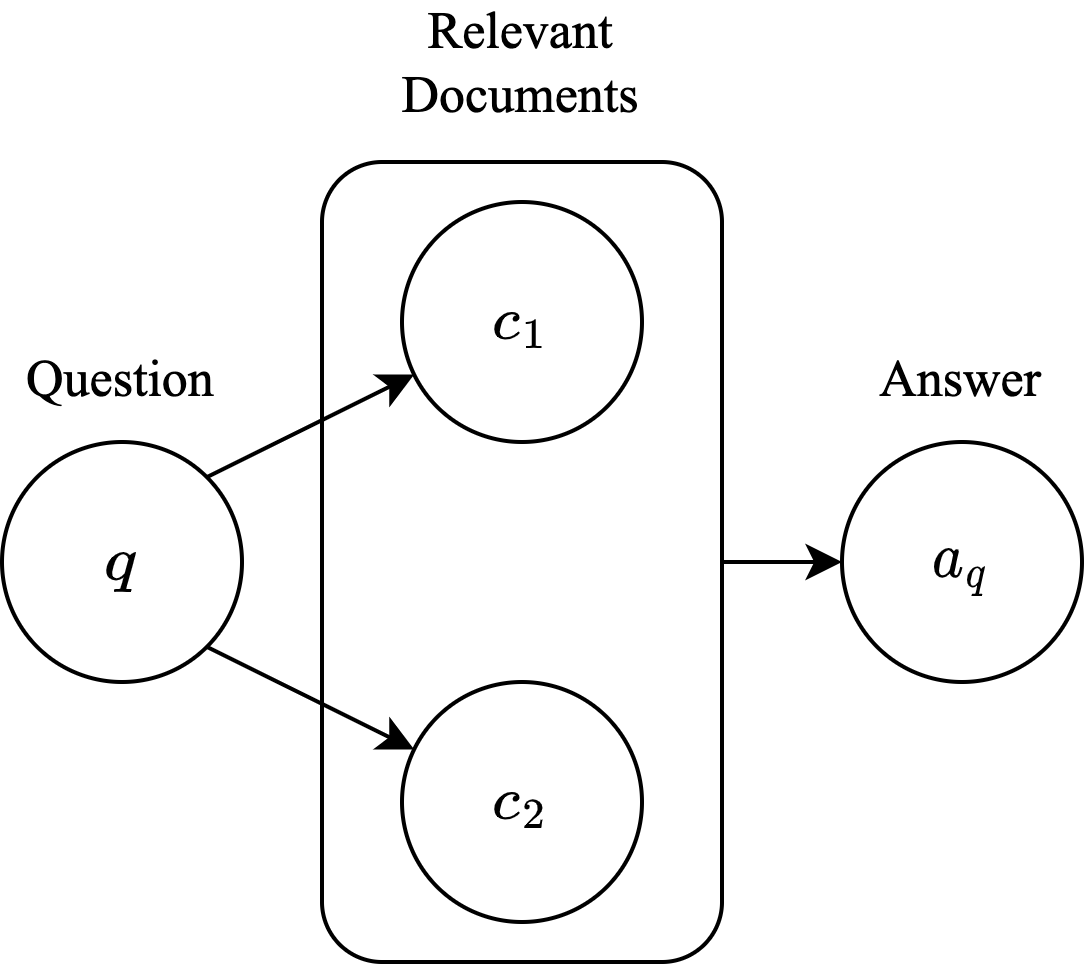}
  \caption{Comparison}
  \label{fig:rpath-cmp}
\end{subfigure}
\hfill
\begin{subfigure}{0.27\textwidth}
\centering\includegraphics[width=\textwidth]{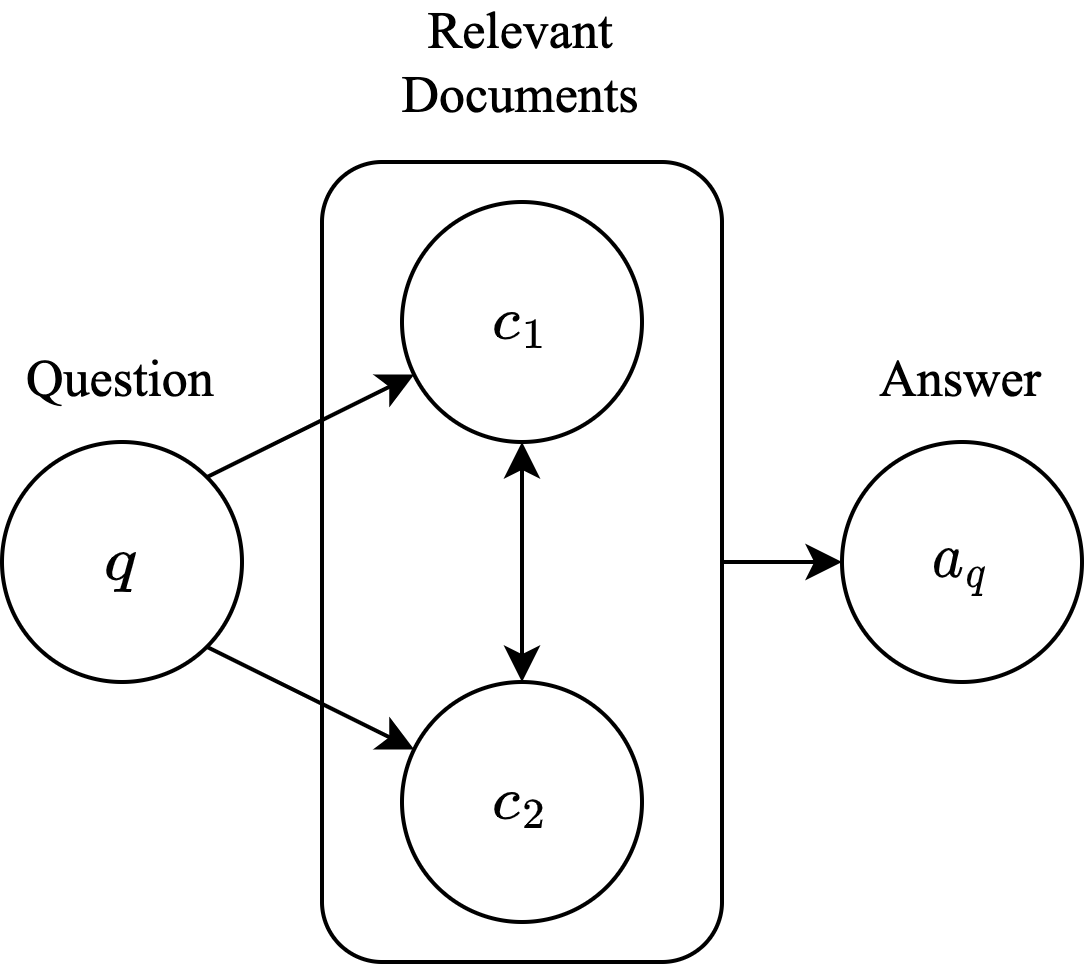}
\caption{Mixed}
\label{fig:rpath-mix}
\end{subfigure}
\caption{Retrieval path types}
\label{fig:rpath}
\end{figure*}

\section{Methodology}
\label{sec:methodology}
Our approach to estimate the performance of a multi-hop question is based on estimating the performance of its hops and combining the estimates. In this section, we present retrieval paths as steps a QA system must go through to gather evidence for a multi-hop question. We then analyze those paths, in terms of the difficulty of retrieving evidence under each path and present our approach for estimating the performance.

\subsection{Retrieval Paths}
For a multi-hop question, one must collect evidence from two or more relevant documents to be able to answer the question. In an open domain setting, those relevant documents are retrieved from a large collection of documents.
A challenge that is unique to multi-hop questions is that the question may not have enough information to retrieve all relevant documents.
Consider a question that must gather information from two supporting documents $d_1$ and $d_2$.
The relationship between those documents and the question can fall into the following cases:

\begin{itemize}
\item There is enough information in the question that allows both supporting documents $d_1$ and $d_2$ to be retrieved. These supporting documents may or may not be closely related.
\item The question has enough information to allow only one supporting document $d_1$ to be retrieved; the other supporting document can be retrieved with the information in both $d_1$ and the question.
\end{itemize}

The \textit{relatedness} between the question and the supporting documents and between two documents may be established syntactically, based on common terms and phrases. The relatedness may also be defined at the semantic level, for example, based on the embedding of the question and those of the documents. The former is the approach taken by a sparse retriever whereas the latter is employed in dense retrievers. For the simplicity of our analysis, we assume the relatedness is defined syntactically.

Consider a graph where each node denotes a document or a question and each edge indicates the relatedness between two documents.
The steps a retriever takes to reach from a question to its answer can be described as a set of paths, all starting from the question and ending with an answer. Figure~\ref{fig:rpath}  shows the set of retrieval paths that can be formed between a question and its two supporting documents.
The path in Figure~\ref{fig:rpath-bridge}, referred to as a \textit{bridge retrieval} path, describes a scenario where the second document cannot be easily retrieved without retrieving the first document. This may describe, for example, a question about an entity that is not explicitly mentioned in the question, but the question provides enough context to retrieve document $d_1$ where the entity name is given.
The path in Figure~\ref{fig:rpath-cmp} dubbed as \textit{comparison retrieval} path, represent a scenario where both supporting documents can be retrieved with the information given in the question, but the reasoning for an answer needs both documents. For example, the question may ask if two persons, each described in a separate document, have the same nationality. It should be noted that, despite the naming, the reasoning can take forms other than a comparison such as an aggregation function. For example, each document may give financial data about a company and the question may ask for the total assets of a parent company that owns two companies. For some questions, the retrieval path type may not be known or easy to detect. This can happen when the two documents are closely related and and they are also closely related to the question text, as depicted in Figure~\ref{fig:rpath-mix}. This is called \textit{mixed retrieval} path meaning that the retriever may consider it as \textit{bridge} or \textit{comparison}. Retrieval paths may be generalized to $n$-hop questions, with $n$ nodes representing the supporting documents and the edges describing possible steps a retriever can take.  

\begin{table*}[]
    \centering
    \begin{tabular}{c p{6in}}
        \hline
        \textbf{Ret. path} & \textbf{Example} \\
        \hline
        Bridge & \textbf{Question}: What year was the actor that co-starred with Sidney Poitier in \textcolor{red}{\textit{Little Nikita}} born? \newline
        \textbf{Context 1:} Little Nikita is a 1988 American cult drama film directed by Richard Benjamin and starring River \textcolor{violet}{\textit{Phoenix}} and Sidney Poitier. The film marks the first collaboration between \textcolor{violet}{\textit{Phoenix}} and Poitier (the second being Sneakers in 1992). \newline
        \textbf{Context 2:} River Jude \textcolor{violet}{\textit{Phoenix}} (born River Jude Bottom; August 23, \textcolor{blue}{\textbf{1970}}, October 31, 1993) was an American actor, musician, and activist. He was the older brother of \textcolor{red}{\textit{Rain Phoenix}}, Joaquin \textcolor{violet}{\textit{Phoenix}}, Liberty \textcolor{violet}{\textit{Phoenix}}, and Summer \textcolor{violet}{\textit{Phoenix}}. \newline
        \textbf{Answer}: \textcolor{blue}{\textbf{1970}}
        \\
        \hline
        Comparison & \textbf{Question}: Were \textcolor{red}{\textit{Stanley Kubrick}} and \textcolor{red}{\textit{Elio Petri}} from different countries? \newline
        \textbf{Context 1}: \textcolor{red}{\textit{Elio Petri}} (29 January 1929 2013 10 November 1982) was an Italian political filmmaker. \newline
        \textbf{Context 2}: \textcolor{red}{\textit{Stanley Kubrick}} (July 26, 1928 March 7, 1999) was an American film director, screenwriter, producer, cinematographer, editor, and photographer. He is ... extensive set designs, and evocative use of music. \newline
        \textbf{Answer}: \textcolor{blue}{\textbf{yes}}
        \\
        \hline
    \end{tabular}
    \caption{Two examples of retrieval paths from the HotpotQA dataset. The named entities mentioned in both the question and the contexts, shown in \textcolor{red}{red}, may assist the retriever in finding the supporting documents. The common entities between the two contexts, shown in \textcolor{violet}{violet entities}, may also help.}
    \label{tab:reasoning_paths_examples}
\end{table*}

\subsection{Retrieval Paths in HotpotQA}
To study the prevalence of retrieval paths, we construct those paths for questions in HotpotQA, one of the largest multi-hop QA datasets that is public.   
For each question in the dataset, the question and its two supporting documents form the nodes of the graph. An edge is added between $d_1$ and $d_2$, deeming them as relevant, if there is a common term between the two documents and the probability of finding that term in an arbitrary document is low (i.e. below a threshold $P_{thr}$).

Similarly, edges are added between $q$ and $d_1$ and between $q$ and $d_2$ if a relevance can be established. The value of $P_{thr}$ may be determined experimentally based on the corpus statistics. If $P_{thr}$ is close to 0, most of the retrieval paths will form incomplete graphs. On the other hand, when $P_{thr}$ is close to 1, there will be edges for more general terms or even stop words and the number of mixed types will increase.

With $P_{thr}$ is set to $0.001$, our study of multi-hop questions in the training set of HotpotQA reveals that about 20\% of the questions demonstrate a bridge retrieval path, whereas the number of questions that show a comparison retrieval path is around 14\%. A majority 63\% of questions show a mixed retrieval path with enough overlapping terms between the questions and their supporting documents. In the development set, our findings indicate that approximately 19\% of the questions exhibit a bridge retrieval path, around 15\% show a comparison retrieval path, and 63\%, display a mixed retrieval path. Clearly, these are some rough estimates, based on our threshold setting of relatedness ($P_{thr}$). Table \ref{tab:reasoning_paths_examples} gives an example of each retrieval path.
For less than 3\% of questions in both training and development sets, no retrieval path could be detected due to the lack of more specific common terms. 
For example, the question \textit{``Who released the 2012 record of Red?''} forms an incomplete graph, because \textit{Red}, the name of an album, is also commonly used as a color.

Our analysis of the dataset also reveals that some questions can be answered with only one supporting document, and they are not really multi-hop. The retrieval paths for these questions show a strong relatedness edge between the question and one supporting document and the answer also appears in the same document. While the second supporting document is related to the question, it is not required for extracting the answer. For example, the question ``Who was known by his stage name Aladin and helped organizations improve their performance as a consultant?'' can be answered by the document with title ``Eenasul Fateh'' and text ``Eenasul Fateh also known by his stage name Aladin, is a bangladeshi-british cultural practitioner, magician, live artist and former international management consultant'' in the content. The second document given for this question is ``Management consulting'' which is not necessary to answer the question.

\subsection{Difficulty Estimation based on Retrieval Paths}
\label{sec:parameter-rpath}
The difficulty of a question is often tied to its ambiguity with respect to the collection being searched, which  may be estimated using a clarity score~\cite{cronen2002predicting} or a coherence score~\cite{he2008using}. However, such scorings ignore the multi-hop structure of questions and the complex relationships that hold between documents retrieved for each hop. In our approach, each multi-hop question is assigned a retrieval path, and the difficulty of the question can be measured by the cost of retrieving the context documents along the path.
The cost here refers to the number of additional documents retrieved.

Let $P(c|q)$ denote the probability of reaching from question $q$ a context document $c$ that is needed to answer the question. The smaller the probability, the larger the number of documents the retriever has to retrieve before finding $c$. When this probability is one, there is enough evidence to reach the context without incurring additional costs. The expected number of documents to be retrieved, or the cost, may be denoted by $1/P(c|q)$. We use the terms contexts and supporting documents interchangeably in this paper.

Now consider a question $q$ associated with a 2-hop \textit{bridge} retrieval path and context documents $c_1$ and $c_2$, as shown in Figure \ref{fig:rpath-bridge}. The probability of retrieving both contexts can be written as
\begin{equation}
    P_{ret} = P(c_1|q) \times P(c_2|q, c_1)
    \label{eq:bridge}
\end{equation}
where $P(c_1|q)$ is the probability of reaching $c_1$ from $q$ and $P(c_2|q, c_1)$ is the probability of reaching $c_2$ from $q$ and $c_1$. Here $c_1$ denotes a context that is directly reachable from $q$ but $c_2$ can be reached only after retrieving $c_1$.

For a \textit{comparison} retrieval path, both contexts can be retrieved independently, and the probability of retrieving both contexts can be expressed as
\begin{equation}
    P_{ret} = P(c_1|q) \times P(c_2|q).
    \label{eq:comp}
\end{equation}

For a \textit{mixed} retrieval path, the retriever has three options: (1) retrieve $c_1$ first and $c_2$ next, (2) retrieve $c_2$ first and $c_1$ next, and (3) retrieve both $c_1$ and $c_2$ independently. It is reasonable to assume that the retriever will take the path with the highest probability (or the least cost), i.e.
\begin{equation}
    \begin{aligned}
    P_{ret} = \max\{P(c_1|q) \times P(c_2|q), \\
    P(c_1|q) \times P(c_2|q, c_1),\\
    P(c_2|q) \times P(c_1|q, c_2)\}.
    \end{aligned}
    \label{eq:mix}
\end{equation}

It is possible that a given question does not follow any of the aforementioned retrieval paths, for example, when the question does not provide enough evidence to efficiently retrieve any of its contexts. These are rare cases though, as reported for the HotpotQA dataset in the previous section. We consider these questions difficult, with $P_{ret} \approx 0$.

Finally, estimating the difficulty of a question is hinged on estimating the model parameters, as addressed next.

\subsection{Estimating the Model Parameters}
\label{sec:parameter-est}
Under a pre-retrieval setting, our probabilities can be estimated based on the question and maybe the corpus statistics. This means we may not have enough information about some of the hops (e.g. the 2nd-hop document in a bridge question). 

\subsubsection{Estimating the probabilities}
Suppose the retrieval path of a question is known, and the goal is to estimate the probabilities of reaching the hops on the path\footnote{In the next subsection, we discuss how retrieval paths can be predicted.}.
A sparse retriever will use the terms of the question to find the context documents, but selecting those terms for each hop of a multi-hop question is not straightforward.

Unlike a single-hop retriever that uses all question terms in the retrieval, a multi-hop retriever may use named entities that are mentioned and their relationships to guide the search \cite{yang-etal-2018-hotpotqa}.
On the same basis, named entities are good candidates for retrieving the supporting documents at each hop.

Sometimes the question has long phrases (e.g. the title of a song) that appear as a whole in context documents, and those phrases may not be detected as named entities. 
\citet{yadegari2022detecting} study those phrases, referred to as frozen phrases, and show that identifying them can improve the retrieval models in open-domain QA. Thus frozen phrases may also be considered.

We utilize publicly available code for extracting named entities\footnote{https://huggingface.co/Jean-Baptiste/roberta-large-ner-english} and frozen phrases\footnote{https://github.com/Aashena/Frozen-Phrases}. However, named entities or frozen phrases may not appear verbatim in the supporting documents, and this will be a problem in calculating the probabilities. For example, consider the question ``Which singer is in the duo Sugarland, Jennifer Nettles or Roger Taylor?'' from the HotpotQA dataset. In the supporting documents, ``Roger Meddows Taylor'' appears instead of ``Roger Taylor.''
To deal with this problem, we extract unigrams, bigrams, and trigrams\footnote{These are consecutive terms forming bigrams and trigrams.} from named entities. We only extract unigrams from frozen phrases because named entities are already extracted by the named entity extraction module, and the remaining terms in frozen phrases may not be consecutive. With this strategy, it is more likely that some n-grams will appear in supporting documents. Let $NG_q$ denote all those n-grams of query $q$.

Now consider questions that follow a comparison retrieval path. A  hypothesis is that such questions are expected to mention two or more entities (see the example given in Table~\ref{tab:reasoning_paths_examples}) and each hop closely relates to one of those entities. Based on this hypothesis, a 2-hop retriever may extract two unique named entities of the question and retrieve the relevant documents of each named entity. 
Our probabilities are also estimated based on those unique named entities and frozen phrases. In particular, we select the two most specific n-grams $n_1,n_2 \in NG_q$ to represent the two contexts of $q$, and estimate $P(c_1|q)=P(c_1|n_1) = \frac{1}{N(n_1)}$ and $P(c_2|q)=P(c_2|n_2) = \frac{1}{N(n_2)}$, where $N(n) \neq 0$ denotes the number of documents that mention n-gram $n$.

For a bridge question, the probabilities may be estimated similarly, with the exception that only the first context $c_1$ can be reached from the question. In particular, the probability of reaching the first context may be estimated under a \textit{Max} scheme, i.e.
\begin{equation}
\label{eq:max_schema}
    P(c_1|q) = max_{t \in NG_q,\\ N(t)>0} \frac{1}{N(t)}.
\end{equation}
This is an optimistic estimation that assumes the most specific n-gram (i.e., the one with the highest probability) appears in the first supporting document.
Since a pre-retrieval method does not have any additional information about the retrieved documents in the second hop, the probability of reaching the second hop in Equations \ref{eq:bridge} and \ref{eq:mix} can be set to a constant (i.e., $P(c_2|q, c_1) = P_{hop_2}$).

\subsubsection{Detecting the retrieval path of questions}
\label{sec:type-detection}
To use our performance prediction models in Eq. \ref{eq:bridge} and \ref{eq:comp}, one must know the retrieval path of questions. Generally, detecting the retrieval path of a question without detailed information about the supporting documents is not easy. However, the structure of a question and the relationships between the entities mentioned often provide clues on the retrieval path the question may take. Based on those features, a classifier may be trained to detect the retrieval path type of a question.

\section{Experimental Evaluation}
\label{sec:results}
\subsection{Datasets}
\noindent \textbf{HotpotQA}~\cite{yang-etal-2018-hotpotqa}, one of the largest public benchmark for multi-hop QA, includes 113k Wikipedia-based question-answer pairs. The dataset is broken down to 90k train set, 7.4k dev set, 7.4k test-distractor where 2 gold paragraphs are mixed with 8 distractors (closed-domain) and 7.4k test-fullwiki where the relevant paragraphs include the first paragraph of all Wikipedia articles (open-domain). The train set is also broken down to 18k train-easy (mostly single-hop), 56.8k train-medium (multi-hop) and 15.7k train-hard (multi-hop hard) questions. The train-easy set is detected by labelling the turkers who tended to type single-hop questions. The train-medium class includes questions that could be answered with high confidence using a QA system built upon Clark and Gardner \cite{clark2018simple} with some of the SOTA techniques added \cite{yang-etal-2018-hotpotqa}. Questions in the train and dev sets are also tagged as either bridge or comparison. 

\noindent\textbf{WikiPassageQA}~\cite{cohen2018wikipassageqa}, is the largest non-factoid open-domain QA dataset including 4k questions created from 863 Wikipedia documents. The dataset consists of train, dev and test sets including 3332, 417, and 416 questions. Each question can be answered with multiple passages from one long document.

\noindent\textbf{WikiQA} \cite{yang-etal-2015-wikiqa} is an open-domain dataset including 3k single-hop questions created from Bing query logs and it broken into train, dev and test sets including 2118, 296, and 633 questions. In WikiQA, each question can be answered with a Wikipedia document.

\subsection{Evaluation Metrics}
\label{sec:experimental_setup}
\subsubsection{Correlation with the average precision}
The quality of our query difficulty estimation may be gauged in terms of \textit{the correlation} between our model performance estimates and the actual performance of the retrievers, as commonly done in ad-hoc and QA retrieval \cite{hauff2008improved, carmel2006makes, arabzadeh2021bert}. We report this metric in terms of Pearson's correlation (P-$\rho$), Spearman's correlation (S-$\rho$), and Kendall's correlation (K-$\tau$). Significant test results at p-values \num{0.01} and \num{0.001} against the null hypothesis that the distributions are uncorrelated are also reported.  

As a measure of the performance of a retriever on a question,  we use the average precision in retrieving documents that are needed to answer the question. However, unlike ad-hoc retrieval where there is one list of retrieved documents, multi-hop QA retrieval involves retrieving supporting documents for multiple hops. To aggregate these results into a single list,  we interleave the documents from different hops, with the first document from the first hop, followed by the first document from the second hop, etc. This strategy, which is also used in \citet{xiong2020answering}, evenly combines the results for different hops and is expected to describe the behaviour of retrievers.

\subsubsection{Pairwise difficulty estimation accuracy}
Pairwise difficulty estimation, which determines which one of two questions is more challenging to evaluate, is a more modern performance metric utilized in recent post-retrieval studies \cite{datta2022deep}. This metric is more intuitive compared to correlation, which can be challenging due to the the disparity between the distribution of the estimated scores and the distribution of the actual evaluation metric, such as average precision. 

We compare the difficulty of question pairs by utilizing the estimated scores obtained from pre-retrieval models. These scores indicate whether $q_1$ is more difficult than $q_2$. Furthermore, by considering the positions of supporting documents in the list of retrieved documents, we can say that $q_1$ is more challenging than $q_2$ if the number of retrieved documents to cover both supporting documents is greater for $q_1$. Hence, the accuracy of a model can be estimated based on the labels it assigns to question pairs and the actual labels of those pairs during evaluation.

\subsubsection{Paragraph exact match and recall}
Question difficulty classes may be defined and each question may be assigned to a class based on a performance prediction model. The actual performance of retrievers on those classes can show how good the prediction model has performed. Our question classes include \textit{easy}, \textit{hard} and \textit{extra hard}. 

The actual performance of a QA system on each class is measured in terms of the fraction of questions whose supporting documents are all retrieved and the fraction of questions that at least one of their supporting documents is found. These performance measures are referred to in the literature~\cite{zhang2021pathranker,xiong2020answering} as  \textit{Paragraph Exact Match (PEM)} and \textit{Paragraph Recall (PR)} respectively.

\subsubsection{Answer exact match and F1-score}
For the end-to-end performance of a QA system, we use Exact Match (EM) and the F1 score, following the prior work on QA evaluation\cite{xiong2020answering, zhang2021pathranker, qi2019answering, yang-etal-2018-hotpotqa}. The former measures if a returned answer exactly matches the ground truth and the latter combines the precision and recall in terms of the number of common words between a predicted answer and the ground truth.

\begin{table*}[]
    \centering
    \begin{tabular}{|c|c|c|c|c|c|c|}
        \hline
         \multirow{2}{*}{QPP Baseline} & \multicolumn{2}{c}{Bridge} & \multicolumn{2}{|c}{Comparison} & \multicolumn{2}{|c|}{Mixed} \\
         \cline{2-7}
        & MDR & GoldEn & MDR & GoldEn & MDR & GoldEn\\
        \hline
        SCS \cite{he2004inferring} & 49.93 & 50.86 & 51.34 & 54.06 & 53.71 & 54.81 \\
        maxSCQ \cite{zhao2008preretrieval} & 53.94 & 54.03 & \textbf{53.85} & 56.83 & 54.67 & 55.26 \\
        avgSCQ \cite{zhao2008preretrieval} & 52.46 & 52.05 & 51.67 & 57.15 & 55.46 & 56.01 \\
        maxIDF \cite{cronen2002predicting} & 53.81 & 53.83 & 53.78 & 57.03 & 54.61 & 55.20 \\
        avgIDF \cite{cronen2002predicting} & 52.44 & 51.99 & 51.52 & 56.99 & 55.41 & 55.91 \\
        maxIEF \cite{arabzadeh2020neural} & 52.11 & 51.19 & 50.56 & 50.92 & 51.08 & 50.50 \\
        avgIEF \cite{arabzadeh2020neural} & 50.23 & 50.21 & 50.37 & 50.07 & 50.21 & 50.15 \\
        multHP (ours) & \textbf{58.82} & \textbf{58.90} & 52.50 & \textbf{57.73} & \textbf{57.39} & \textbf{58.06} \\
        \hline
    \end{tabular}
    \caption{Pairwise difficulty estimation accuracy compared to pre-retrieval QPP baselines}
    \label{tab:accuracy_baselines}
\end{table*}

\begin{table*}[]
    \centering
    \begin{tabular}{|l|l|l|l|l||l|l|l|}
         \hline
         \textbf{Question} & \multirow{2}{*}{\textbf{QPP Baseline}} & \multicolumn{3}{c||}{\textbf{MDR}} & \multicolumn{3}{c|}{\textbf{GoldEn}} \\
         \cline{3-8}
         \textbf{Type} & & P-$\rho$ & S-$\rho$ & K-$\tau$ & P-$\rho$ & S-$\rho$ & K-$\tau$ \\
         \hline
         \hline
         \multirow{4}{*}{Bridge} & maxSCQ \cite{zhao2008preretrieval} & 0.1418 & 0.1263 & 0.0907 & 0.1477 & 0.1479 & 0.1081 \\
         & maxIDF \cite{cronen2002predicting} & 0.1405 & 0.1259 & 0.0904 & 0.1431 & 0.1450 & 0.1061 \\
         & maxIEF \cite{arabzadeh2020neural} & 0.0266 & 0.0304 & 0.0220 & 0.0164 & 0.0067 & 0.0050 \\
         & multHP (ours) & \textbf{0.2342} & \textbf{0.2480} & \textbf{0.1849} & \textbf{0.2858} & \textbf{0.3088} & \textbf{0.2369} \\
         \hline
        \multirow{4}{*}{Comparison}
         & maxSCQ \cite{zhao2008preretrieval} & \textbf{0.0866} & 0.1051 & 0.0829 & 0.1794 & 0.1977 & 0.1548 \\
         & maxIDF \cite{cronen2002predicting} & 0.0783 & 0.0941 & 0.0742 & \textbf{0.1923} & 0.2050 & 0.1604 \\
         & maxIEF \cite{arabzadeh2020neural} & -0.0195 & -0.0076 & -0.0055 & 0.0342 & 0.0070 & 0.0057 \\
         & multHP (ours) & 0.0460 & \textbf{0.1139} & \textbf{0.0894} & 0.1130 & \textbf{0.2597} & \textbf{0.2024} \\
         \hline
    \end{tabular}
    \caption{Correlation between the difficulty prediction of pre-retrieval models and the actual retriever performance, in terms of average precision, of MDR and GoldEn on HotpotQA (results are statistically significance at p-value < 0.001)}
    \label{tab:corr_mdr_golden}
\end{table*}

\subsection{Retrieval Models and QPP Baselines}
In the absence of prior research on QPP in multi-hop QA settings, we use the following pre-retrieval methods commonly used in ad-hoc retrieval tasks as baselines for comparison: (1) Inverse Document Frequency (IDF)~\cite{cronen2002predicting, scholer2004query}, which predicts query performance by considering the specificity of the question terms, with higher values indicating an easier question to answer; 
(2) Simplified Clarity Score (SCS)~\cite{he2008using}, which estimates query performance by taking into account both the query length and the specificity by computing the divergence between the simplified query language model and the collection language model; 
(3) Collection Query Similarity (SCQ)~\cite{zhao2008preretrieval}, which predicts the performance based on the similarity between the query and the collection documents; 
(4) Inverse Edge Frequency (IEF) \cite{arabzadeh2020neural}, which estimates question specificity within an embedding space, taking into account the number of close neighbors associated with each term. These predictors are aggregated over question terms using max and average functions, resulting in maxIEF, avgIEF, maxIDF, avgIDF, maxSCQ, avgSCQ, and SCS.

With the performance of open-domain QA systems typically bounded by the retrievers \cite{lee2018ranking, nie-etal-2019-revealing}, query difficulty may be quantified in terms of the performance of the retrievers. We utilize two multi-hop and one single-hop QA retrievers in our evaluation. \textbf{GoldEn}~\cite{qi2019answering} is a sparse model built on top of DrQA~\cite{chen-etal-2017-reading}. We used the authors public code \footnote{https://github.com/qipeng/golden-retriever} and instructions to train the retriever on the HotpotQA dataset. \textbf{MDR}~\cite{xiong2020answering} is a dense model that retrieves the relevant documents based on the inner product score between a question and documents embedding vectors. In our work, we used the public code \footnote{https://github.com/facebookresearch/multihop\_dense\_retrieval} and retriever checkpoint provided by the authors. We utilized the same hyperparameters as reported in the original study, with the exception of the number of retrieved documents, which we set to 5 per hop. We also used DrQA~\cite{chen-etal-2017-reading} as a single-hop retriever.

\subsection{Compared to Pre-retrieval QPP Baselines}
\label{sec:ranked_based_analysis}
\subsubsection{Pairwise Difficulty Comparison}
Table \ref{tab:accuracy_baselines} shows the accuracy of correctly predicting the pairwise difficulty of questions. The results are categorized based on the question types introduced in Section \ref{sec:parameter-rpath}. There are 5,918 Bridge questions (17,508,403 question pairs), and our metric outperforms all of the baselines with over 4.87\% improvement (corresponding to more than 852k question pairs) in both retrievers. In Comparison questions, our multHP slightly improves the accuracy, while our estimate for MDR's performance does not exceed the baselines. Based on our analysis, MDR performs quite well in retrieving both supportive documents, and our predicted scores for questions may not accurately estimate the retriever's performance. In addition to Bridge and Comparison questions, we leverage Equation \ref{eq:mix} to demonstrate the effectiveness of our proposed formulations even when the question type is not specified. In the Mixed type, our multHP outperformed all of the baselines with improvements of 1.93\% and 2.05\% in terms of accuracy for the MDR and GoldEn models, respectively.

\subsubsection{Multi-hop Pointwise Correlation}
Table~\ref{tab:corr_mdr_golden} shows the correlation between the predicted query performance scores and the actual performance of the  retrievers.  The results are broken down to question types, in terms of comparison or bridge, since the calculations are slightly different and the retrieval path of a question can be easily predicted.

Our predicted scores showed a significant correlation with the actual performance of the both retrievers for bridge questions.

However, the Pearson correlation for comparison questions was lower for MDR \cite{xiong2020answering} simply because MDR performs quite well in comparison questions due to the fact that both entities are mentioned in the questions. This results in both supporting documents being placed in the low ranks of the retrieved results for either one hop or both, and an estimation solely based on a random document selection model may not correlate with the actual average precision. Ultimately, in comparison questions, neither our estimated scores nor those of our baseline methods show strong correlation with the actual performance of the MDR retriever.  Our analysis of five retrieved documents showed that the average ranks of the two supporting documents that appeared in the 1st hop were 1.71 and 1.72 for MDR, and those ranks were 1.63 and 4.52 for GoldEn. The difference in the average ranking of the second document in the hop 1 indicates that MDR performs exceptionally well on these questions. Besides, considering the disparity in scale between our estimated scores and the average precision, particularly for comparison questions, Spearman and Kendall coefficients may serve as better indicators of performance due to their rank-based nature.

Comparing the results of syntactic metrics such as multHP and SCQ with the IEF semantic metric \cite{arabzadeh2020neural} shows that the mere semantics of terms may not be a good estimator of specificity. One of the main challenges of embedding-based metrics is out-of-vocabulary terms, which which include many named entities. Most multi-hop questions are factoid questions that include named entities, and these named entities may not map well to the embedding space. Furthermore, IEF merely leverages the embedding space and ignores the corpus statistics, which are good predictors for retrieval performance prediction. 

\begin{table*}[]
    \centering
    \begin{tabular}{|l|l|l|l|l|l|l|}
        \hline
        \multirow{2}{*}{\textbf{QPP Model}} & \multicolumn{3}{c|}{\textbf{WikiPassageQA}} & \multicolumn{3}{c|}{\textbf{WikiQA}} \\
        \cline{2-7}
        & P-$\rho$ & S-$\rho$ & K-$\tau$ & P-$\rho$ & S-$\rho$ & K-$\tau$ \\
        \hline
        maxSCQ \cite{zhao2008preretrieval} & 0.1107 & 0.1363 $\ast$ & 0.0980 $\ast$ & -0.0015 & -0.0230 & -0.0168\\
        maxIDF \cite{cronen2002predicting} & 0.1096 & 0.1287 $\ast$ & 0.0929 $\ast$ & -0.0402 & -0.0432 & -0.0321 \\
        maxIEF \cite{arabzadeh2020neural} & 0.1164 & 0.0823 & 0.0621 & -0.0816 & -0.0801 & -0.0602 \\
        multHP (ours) & \textbf{0.1889} $\dagger$ & \textbf{0.2507} $\dagger$ & \textbf{0.1833} $\dagger$ & \textbf{0.1030} $\ast$ & \textbf{0.1280} $\ast$ & \textbf{0.0981} $\ast$ \\
        \hline
    \end{tabular}
    \caption{Correlation between the difficulty prediction of QPP models and the actual retriever performance, in terms of average precision, of DrQA on WikiPassageQA and WikiQA datasets ($\ast$ and $\dagger$ denote the correlations with p-value less than 0.01 and 0.001 respectively)}
    \label{tab:corr_single_hop}
\end{table*}

The same results was observed using GoldEn \cite{qi2019answering}, a sparse retriever. As shown in Table \ref{tab:corr_mdr_golden}, the correlation between our estimated difficulty score and the actual performance of the GoldEn retriever was much more pronounced, compared to the baselines, for bridge questions. However, for comparison questions, our approach did not perform better than the baselines. Our error analysis shows that the GoldEn retriever could not find the supporting documents for questions that we estimated as easy questions with a fairly high score. For instance, consider the question ``Which pizza shop opened first, Toppers Pizza or America's Incredible Pizza Company?''. Based on our probabilities, $P(c_1|Toppers\ Pizza) = 0.33$ and $P(c_2|America's\ Incredible) = 1$, but the GoldEn retriever failed to retrieve the supporting document ``America's Incredible Pizza Company''. In this particular case, GoldEn query generator emitted two queries, ``pizza shop opened first, Toppers Pizza'' and ``Pizza,'' to retrieve the supporting documents. These queries failed to retrieve the second entity. Also in some cases, the GoldEn retriever could extract supporting documents for questions that we estimated as difficult to retrieve. Consider the comparison question ``Hayden is a singer-songwriter from Canada, but where does Buck-Tick hail from?''. Our approach correctly extracted the two entities, computed the probabilities $P(c_1|Buck-Tick) = 0.0164$ and $P(c_2|Hayden) = 0.0011$, and estimated the $P_{ret} \approx 1.8 \times 10^{-5}$. However, Golden retriever extracted both supporting documents easily at top of the result set using the two queries ``Hayden is a singer-songwriter from Canada'' and ``Buck-Tick''. In this case, we underestimated the probability of finding a relevant document by considering only one entity per query while the retriever leveraged all information in the query, such as ``singer-songwriter'' and ``Canada'' in this example.

\subsubsection{Single-hop Pointwise Correlation}
To show how our model performs on single-hop questions, we evaluated our approach on the test sets of WikiPassageQA \cite{cohen2018wikipassageqa} and WikiQA \cite{yang-etal-2015-wikiqa}, two open domain QA datasets using DrQA retriever, following the setting as explained in Section \ref{sec:ranked_based_analysis}.
We used the setting of our bridge questions where the estimation is done based on the first hop.
From Table~\ref{tab:corr_single_hop}, we can observe that while the correlations are not very strong, our estimates show a stronger correlation with the actual performance of the system, compared to the baselines, and the results are statistically significant. This is mostly because of our term selection strategies and the use of named entities and frozen phrases for our performance prediction.

\subsection{Performance Across Difficulty Classes}
\subsubsection{Retriever performance results}
In another experiment, we wanted 
to illustrate the performance drop in different difficulty classes. To this aim, we categorized the questions of HotpotQA's dev set into different difficulty classes and evaluated the performance of the retrievers on those class. A similar categorization is done in the work of \citet{mothe2019impact}. To set threshold scores for categorizing questions into different difficulty classes, we used the percentile-based strategy \cite{mothe2019impact}. For bridge questions, we set $P_{hop_2} = 0.125$ to calibrate the score difference between bridge and comparison questions. We named these difficulty classes extra hard (1st quartile), hard  (2nd quartile), easy  (3rd and 4th quartiles). We merged 3rd and 4th quartiles into one class because we observed that there was no noticeable difficulty gap between these two sets. 

Table \ref{tab:retriever_max} shows the performance of three retrievers, with their default settings, on different question classes in terms of PEM and PR under \textit{Max} scheme. We can see more than 10\%, 16\%, and 15\% performance drop between easy and extra hard classes of DrQA, GoldEn, and MDR respectively in terms of PEM. Since both supporting documents are required to answer a multi-hop question, by comparing the results of the retrievers on question classes, we can conclude that the number of questions that cannot be answered in hard and extra hard classes are considerably larger than the number of such questions in the easy class.

\begin{table*}[]
    \centering
    \begin{tabular}{|l|c|c|c|c|c|c|c|c|c|c|}
        \hline
         \multirow{3}{*}{\textbf{Class}} & \multicolumn{10}{c|}{\textbf{Model}}  \\
         \cline{2-11}
         & \multicolumn{2}{c|}{DrQA (k=100)} & \multicolumn{4}{c|}{GoldEn (k=5)} & \multicolumn{4}{c|}{MDR(k=1)} \\
         \cline{2-11}
         & PEM & PR & PEM & PR & EM$_{joint}$ & F1$_{joint}$ & PEM & PR & EM$_{joint}$ & F1$_{joint}$ \\
        \hline
         Easy & 33.07 & 84.70 & 67.01 & 95.78 & 20.68 & 43.87 & 71.50 & 89.97 & 36.86 & 59.66 \\
         Hard & 27.49 & 77.27 & 60.31 & 88.63 & 20.25 & 39.31 & 63.94 & 83.59 & 35.62 & 55.68 \\
         Ex. Hard & 22.69 & 71.45 & 51.61 & 80.09 & 16.95 & 35.27 & 56.71 & 73.65 & 31.49 & 48.06 \\
         \hline
    \end{tabular}
    \caption{Retrieval performance (in terms of PEM and PR) and end-to-end performance (in terms of EM and F1) of three models across three difficulty classes, predicted using the \textit{Max} scheme, showcasing performance degradation for more challenging questions}
    \label{tab:retriever_max}
\end{table*}

\subsubsection{End-to-end performance results}
Our analysis so far shows that our QPP model is quite effective in predicting the performance of multi-hop QA retrievers. In this section, we want to evaluate how this translates to predicting the end-to-end performance of our QA systems on different difficulty classes. To evaluate the performance in both document retrieval and answer extraction phases, we used the standard answer and supporting facts given in the dataset, and calculated Exact Match (EM) and F1 score on questions with different classes of difficulty. Tables \ref{tab:retriever_max} shows the results of GoldEn and MDR models, in an end-to-end setting, in terms of joint EM and F1 score for answer extraction and supporting sentences prediction. We can observe a declining performance between easy and extra hard classes. It should be noted that our difficulty score estimation is merely based on the retrieval paths, and some questions that are deemed difficult to answer, using the QPP model, may not pose much challenge to the QA system in finding the supporting documents. 

\subsection{Two Use Cases}
\subsubsection{Dataset annotation}
Annotating datasets such as HotpotQA can help in evaluating the models, and our performance prediction can be used in the annotation process. In particular, detecting question difficulty prior to the retrieval phase may allow, for example, the models to be evaluated on more difficult subsets. Or one may choose simpler models for easy questions and more complex models to answer difficult questions.  The train and dev sets in HotpotQA have annotations for the question difficulty and the type of retrieval path, but those annotations are not given for the test set.

The test set of HotpotQA does not have question types (e.g. comparison or bridge). For our experiments on the test set, we trained a type detection model, as discussed in Section~\ref{sec:type-detection}. To train the model, we used 80\% of the HotpotQA train set for training and the other 20\% as the development set because the test set does not have retrieval path types of questions. For training, we utilized the [CLS] embedding created by RoBERTa$_{base}$ followed by a hidden layer with size 128 to extract high-level features. Finally, the last layer is a softmax layer to predict the probability of each type (i.e., bridge or comparison). We used \textit{relu} activation function and set learning rate $5e-5$, batch size 64, and epoch 3. We evaluated this model using the actual dev set of HotpotQA, and its performance was 99.63\%, 94.64\%, and 97.07\% in terms of precision, recall, and F1-score respectively.

\subsubsection{Adaptive retrievers}
\begin{figure}
    \centering
    \includegraphics[width=\linewidth]{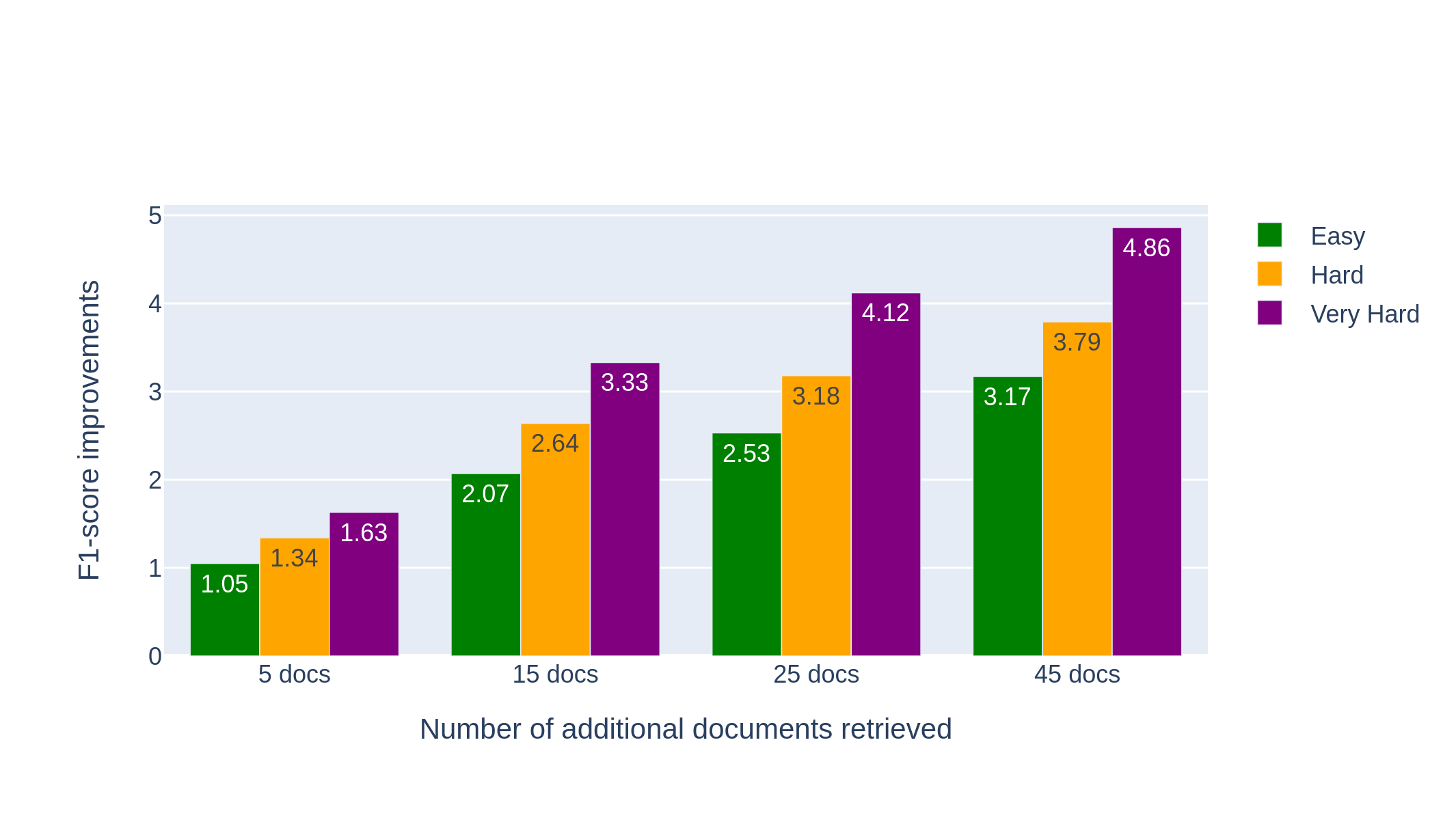}
    \caption{Improvements in the end-to-end performance of MDR \cite{xiong2020answering}, in terms of F1-score, across different difficulty classes and varying the  number of additional document retrieved, showing larger improvements for more difficult classes}
    \label{fig:f1_additional_docs_mdr}
\end{figure}

Improving the performance of models is possible by detecting the difficulty levels of questions and maybe allocating additional resources for more difficult questions. Figure \ref{fig:f1_additional_docs_mdr} shows how retrieving additional documents improves the end-to-end performance of MDR model. We can observe that increasing the number of retrieved documents have most positive impact on the extra hard set compared to the easy set.

\begin{figure}
    \centering
    \includegraphics[width=\linewidth]{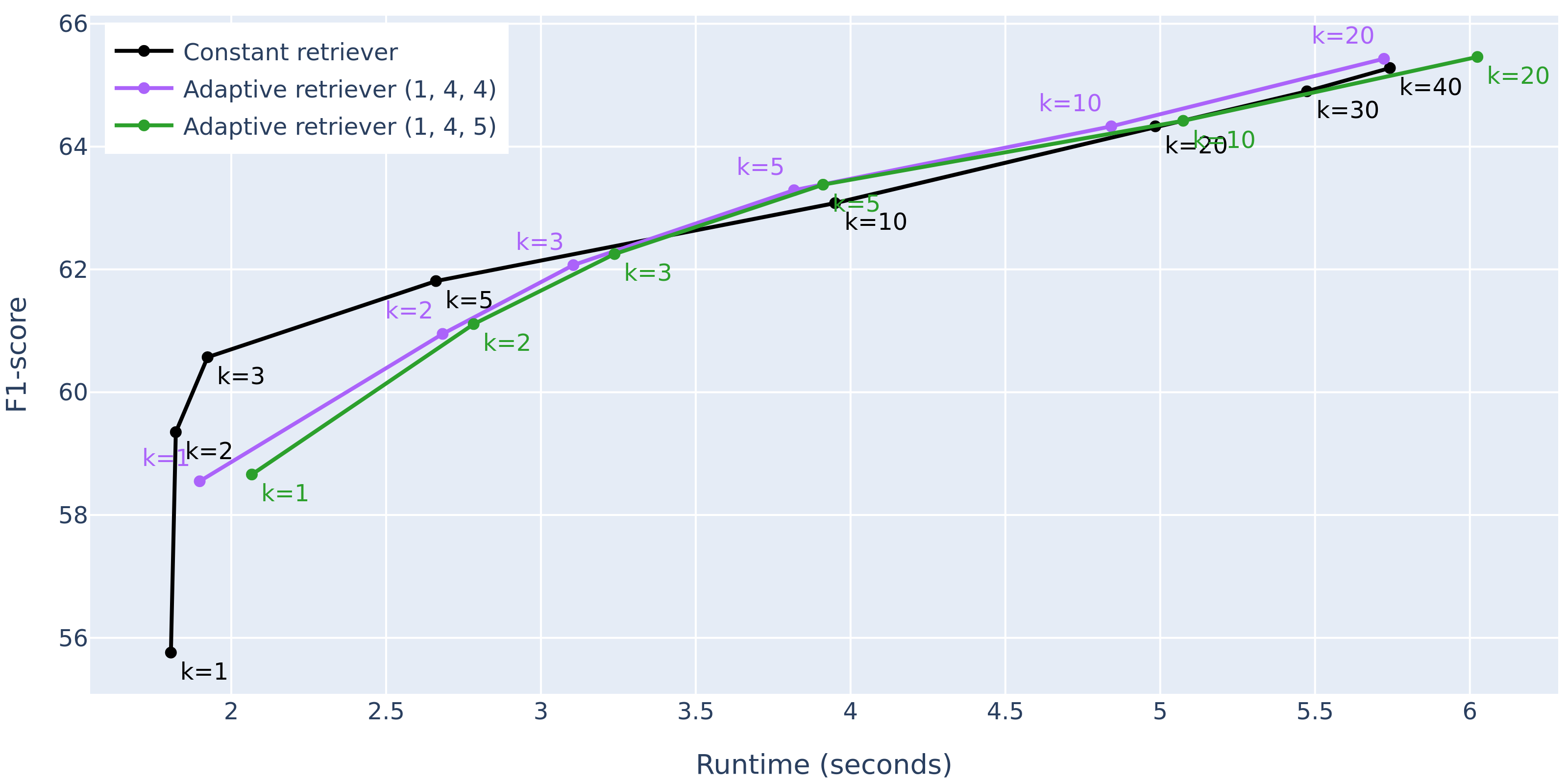}
    \caption{Performance, in terms of F1-score, of MDR \cite{xiong2020answering} with the adaptive retriever compared to a constant retriever while k varied, showing that the adaptive retriever achieves a higher performance under the same budget}
    \label{fig:f1_runtime}
\end{figure}

To take another step forward, we built an adaptive retriever that retrieved more passages for questions that were detected to be difficult. In particular, the retriever fed $ck$ documents to the reader where $c$ was set based on the difficulty class and $k$ varied from \num{1} to \num{20}. Figure \ref{fig:f1_runtime} shows the results with $c$ set to \num{1} for easy, \num{4} for hard and \num{4} and \num{5} for extra hard. Given a limited running time budget, a higher performance, in terms of F1-score, is achieved using the adaptive retriever compared to a constant one that retrieves the same number of documents for all questions irrespective of their difficulty classes. This means the adaptive retriever saves time on easy questions by retrieving fewer documents and spends the saved time on retrieving more documents for difficult questions, which improve the performance of the model. 
We can also see a sharp increase in the performance of the constant retriever as $k$ is increased from \num{1} to \num{3}, indicating that all questions including easy ones benefit from larger $k$ values. However, for $k \geq 5$, easy questions do not benefit as much as hard questions. One possible explanation is that top-ranking documents tend to contain relevant information for easier questions more often than for harder ones. We will release the code and all annotations at GitHub\footnote{https://github.com/MhmDSmdi/performance-prediction-for-multihop-QA}.

\section{Conclusions}
\label{sec:conclusions}
In this paper, we introduce the task of query performance prediction for multi-hop questions. We present an approach to estimate a difficulty score of a multi-hop question based on the clues in the question. We propose retrieval paths based on overlapping terms between the question and its supporting documents. Our experimental evaluation shows significant correlations between the performance of the retrievers used in our evaluation and our estimated difficulty scores, and those correlations are much higher than those obtained by our QPP baselines from the literature. The same trend is observed for the end-to-end models with the performance considerably dropped for the questions that are deemed difficult by our model. Determining the difficulty of a multi-hop question using a pre-retrieval method can assist the retrievers to have a better chance of retrieving all required documents to answer the question. As a possible future direction, our models may be improved by considering more refined retrieval path types and maybe parameter settings. Also analyzing the performance of down-stream tasks using our difficulty score estimation is another possible direction.

\bibliographystyle{ACM-Reference-Format}
\bibliography{references}

\appendix

\end{document}